\documentclass[letterpaper]{article} % DO NOT CHANGE THIS
\usepackage{aaai24}  % DO NOT CHANGE THIS
\usepackage{times}  % DO NOT CHANGE THIS
\usepackage{helvet}  % DO NOT CHANGE THIS
\usepackage{courier}  % DO NOT CHANGE THIS
\usepackage[hyphens]{url}  % DO NOT CHANGE THIS
\usepackage{graphicx} % DO NOT CHANGE THIS
\usepackage{natbib}  % DO NOT CHANGE THIS AND DO NOT ADD ANY OPTIONS TO IT
\usepackage{caption} % DO NOT CHANGE THIS AND DO NOT ADD ANY OPTIONS TO IT
\usepackage{algorithm}
\usepackage{algorithmic}
\usepackage{amsmath}
\usepackage{amssymb}
\usepackage{booktabs}
\usepackage{newfloat}
\usepackage{listings}
\DeclareCaptionStyle{ruled}{labelfont=normalfont,labelsep=colon,strut=off} % DO NOT CHANGE THIS
\floatstyle{ruled}
\newfloat{listing}{tb}{lst}{}
\floatname{listing}{Listing}
\title{Lifting by Image - Leveraging Image Cues for Accurate 3D Human Pose Estimation}
\author {
    Feng Zhou\textsuperscript{\rm 1},
    Jianqin Yin\textsuperscript{\rm 1}\thanks{Corresponding author},
    Peiyang Li\textsuperscript{\rm 1}
}
\affiliations {
    \textsuperscript{\rm 1}School of Artificial Intelligence, Beijing University of Posts and Telecommunications, China\\
    \{zhoufeng,\ jqyin,\ lipeiyang\}@bupt.edu.cn
}
\usepackage{bibentry}
\begin{document}

\maketitle

\begin{abstract}
The ``lifting from 2D pose" method has been the dominant approach to 3D Human Pose Estimation (3DHPE) due to the powerful visual analysis ability of 2D pose estimators. Widely known, there exists a depth ambiguity problem when estimating solely from 2D pose, where one 2D pose can be mapped to multiple 3D poses. Intuitively, the rich semantic and texture information in images can contribute to a more accurate ``lifting" procedure. Yet, existing research encounters two primary challenges. Firstly, the distribution of image data in 3D motion capture datasets is too narrow because of the laboratorial environment, which leads to poor generalization ability of methods trained with image information. Secondly, effective strategies for leveraging image information are lacking. In this paper, we give new insight into the cause of poor generalization problems and the effectiveness of image features. Based on that, we propose an advanced framework. Specifically, the framework consists of two stages. First, we enable the keypoints to query and select the beneficial features from all image patches. To reduce the keypoints attention to inconsequential background features, we design a novel Pose-guided Transformer Layer, which adaptively limits the updates to unimportant image patches. Then, through a designed Adaptive Feature Selection Module, we prune less significant image patches from the feature map. In the second stage, we allow the keypoints to further emphasize the retained critical image features. This progressive learning approach prevents further training on insignificant image features. Experimental results show that our model achieves state-of-the-art performance on both the Human3.6M dataset and the MPI-INF-3DHP dataset.
% Code is available at \url{https://github.com/zss171999645/LiftingByImage}.
\end{abstract}
\section{Introduction}
Monocular 3D Human Pose Estimation (3DHPE) aims to estimate the relative 3D coordinates of human joints from an image. It is a fundamental computer vision task related to a wide range of applications, including human motion forecasting \cite{ ding2022towards, liu2020trajectorycnn}, human action recognition \cite{dang2020dwnet}, human-centric generation \cite{cao2023concept, cao2022lsap, cao2023decreases} and so on.

\begin{figure}[t]
  \centering
%   \fbox{\rule{0pt}{2in} \rule{0.95\columnwidth}{0pt}}
  \includegraphics[width=\linewidth]{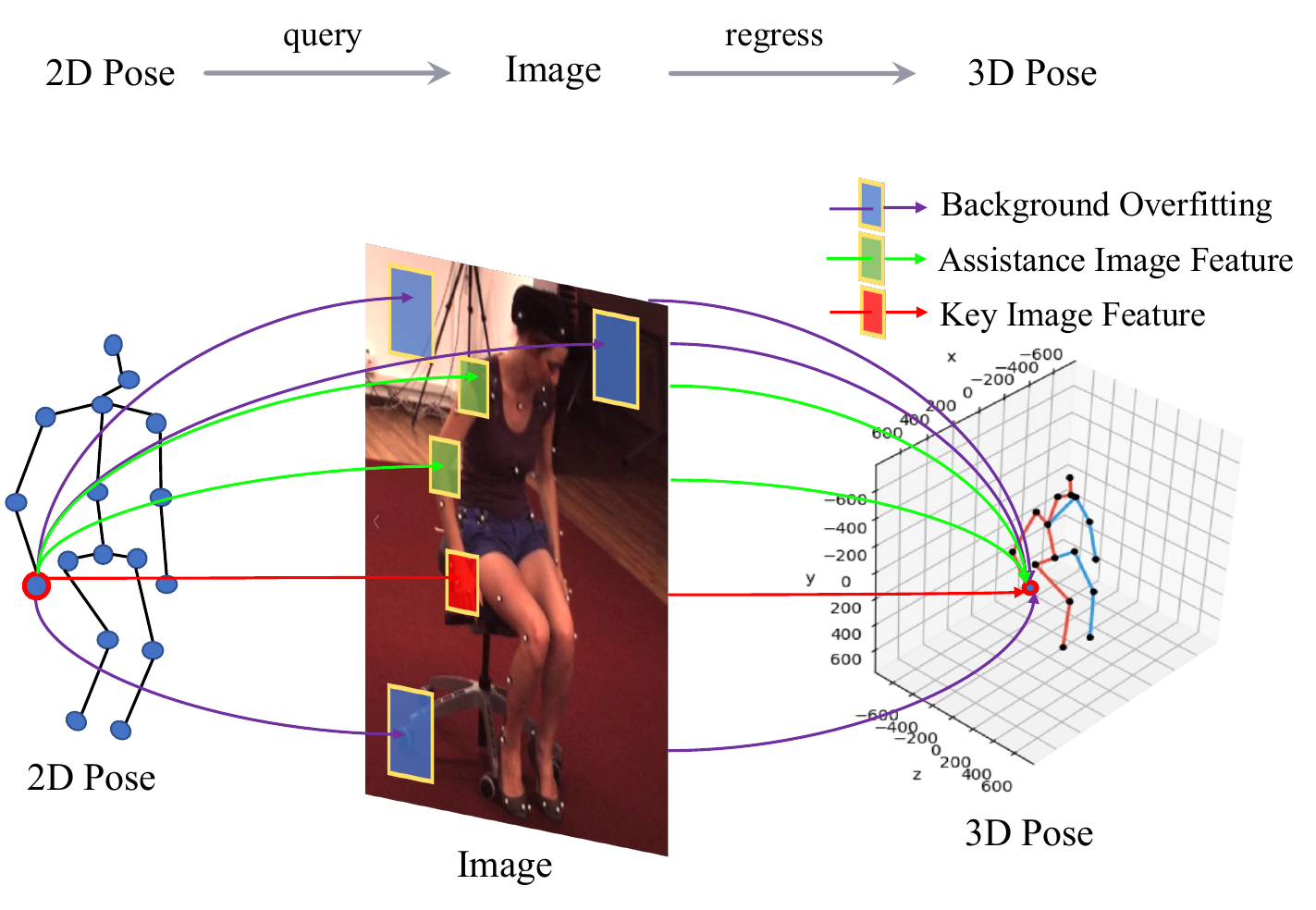}
  \caption{The main idea of this paper is to design a framework that enables 2D poses to regress 3D poses by querying information from the image. The framework is specifically designed based on two key insights: First, excessive attention to dataset-biased background information leads to poor generalization ability. Second, it is not only the image features corresponding to the keypoints that are helpful for the task, but also the associated body structural positions of the keypoints that can provide valuable assistance.}
  \label{fig:intro1}
\end{figure} 
In recent years, 3D human pose estimation has been dominated by the ``lifting" technique \cite{martinez2017simple}. This approach consists of two stages. First, utilize off-the-shelf 2D pose estimators \cite{sun2019deep, newell2016stacked, dang2022learning} to estimate the 2D pose from the image and then regress the 3D pose from the obtained 2D human pose. Compared to direct estimation, this cascaded approach has the following advantages: 2D estimator is trained on more diverse and extensive 2D human pose datasets, which enables stronger visual perception and generalization ability \cite{martinez2017simple}. Besides, the ``lifting" can be trained with infinite 2D-3D pairs by setting different camera views \cite{xu2021monocular}. Nevertheless, estimating 3D pose from 2D pose introduces the depth ambiguity problem, one 2D pose can be mapped to multiple 3D poses. 

Intuitively, rich texture and semantic information in images can assist in regressing a more accurate 3D pose from 2D pose.
\begin{figure}[t]
  \centering
%   \fbox{\rule{0pt}{2in} \rule{0.95\columnwidth}{0pt}}
  \includegraphics[width=\linewidth]{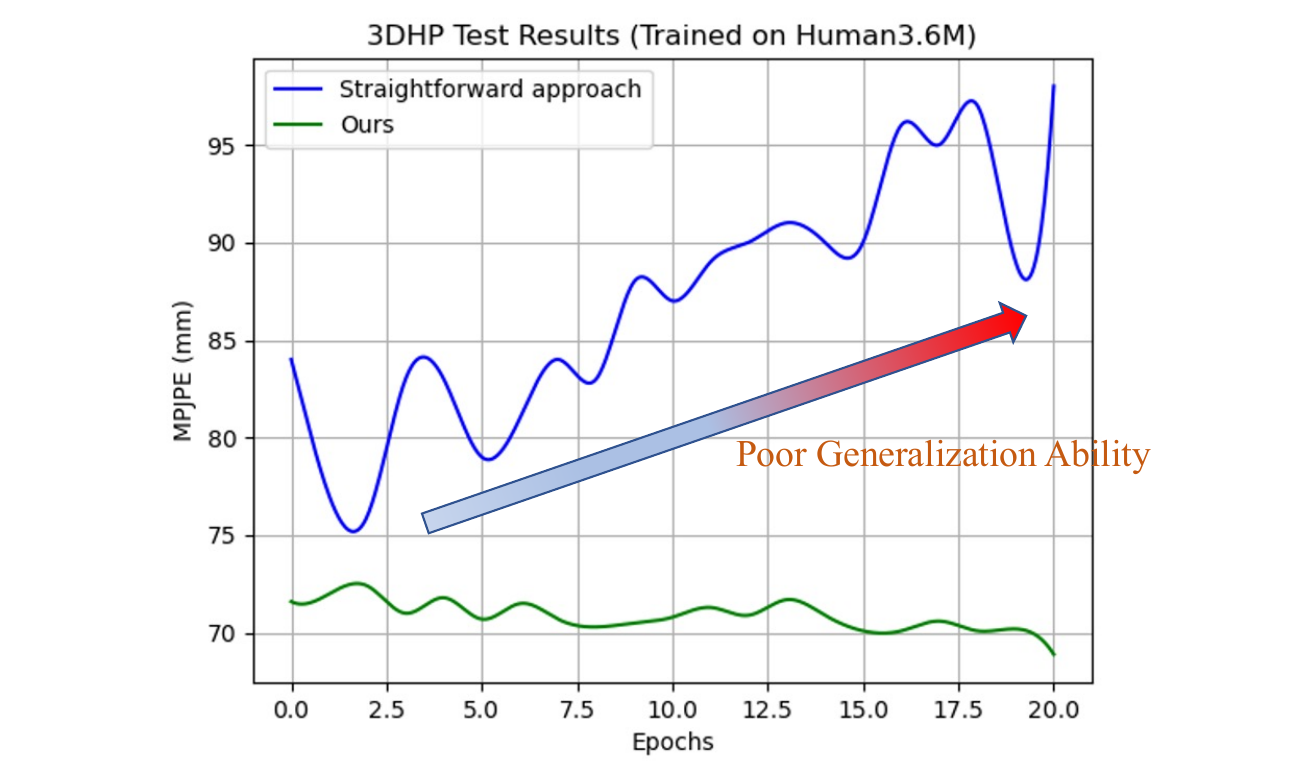}
  \caption{Cross-dataset evaluation between straightforward image-based model and our approach. With continuous training on Human3.6m, the former's accuracy on 3DHP decreased, highlighting poor generalization ability.}
  \label{fig:intro2}
\end{figure} 
There has been some exploration in this direction. For example, \citeauthor{nie2017monocular, xu2021monocular} segment image patches around keypoint locations to aid in generating the 3D pose. Likewise, \citeauthor{zhao2019semantic, liu2019feature} introduced a method of superimposing extracted image features around keypoints' position onto 2D keypoints to offer complementary information to the network. Yet despite the considerable progress, there exist some issues that need to be addressed. Firstly, because 3D human motion datasets were primarily captured in constrained laboratory environments, the distribution of image data is limited. Consequently, methods that are trained with image information tend to suffer from poor generalization ability, as shown in Fig \ref{fig:intro2}. Additionally, effective strategies for leveraging image information are lacking. 

This paper gives a novel insight into the cause of weak generalization ability and the specific effectiveness of image information in predicting the 3D pose. Based on that, we propose a novel framework that estimates 3D human pose from 2D pose by leveraging effective image cues, as shown in Fig \ref{fig:intro1}.

To begin, we utilize the attention mechanism \cite{vaswani2017attention} to study the response of human keypoints to the image features. By analyzing the attention maps, we derived two noteworthy insights: 1. In general, for all keypoints, the attention maps exhibit a high-proportion, wide-range, and indiscriminate emphasis on irrelevant background information outside the human body. This may shed light on the weak generalization abilities of image-based methods, as they overly focus on dataset-biased information. 2. For a specific keypoint, its required image features are not confined solely to its own location in the image. Instead, the required positions also encompass body structure positions that provide depth information for that keypoint. For instance, the features of \emph{elbow} can be instrumental in estimating the depth of \emph{wrist} keypoint. This underscores the constraints of previous methods that exclusively concatenate localized image patches or features around keypoints \cite{nie2017monocular, zhao2019semantic, liu2019feature}.

Based on these understandings, we propose a novel 3D pose estimation framework. The key concept is to allow the keypoints to adaptively focus on critical image features. To give an overview, the progressive learning framework consists of two stages. We enable the keypoints to query and select the beneficial features from all image patches in the first stage, called ``Broad Query". Then we prune the irrelevant image features (mostly background features). At last, we allow the keypoints to further explore information from these critical image features to obtain an accurate 3D pose, called ``Focused Exploration".

Specifically, in Stage 1, we introduce a Pose-guided Transformer Layer, which effectively reduces the keypoints' attention to the background. It leverages the pose-to-image attention matrix to allow the image features to reversely query and aggregate the keypoints features. Through our design, the more crucial image features can extract more relevant information from the keypoint, while less important image features, like background patches, receive comparatively less information. Then we proposed an Adaptive Feature Selection Module, which aims to rank and prune the less important image features by the attention mechanism. In Stage 2, the keypoints are allowed to refocus on critical human image features by several Transformer Layers. Through this cascaded approach, the keypoints are empowered to dynamically explore critical features broadly and simultaneously prevent over-training to the background features. 

We demonstrate quantitative results by conducting our method on standard 3D human pose benchmarks. Experimental results show that our method outperforms state-of-the-art performance on Human3.6M \cite{ionescu2013human3} and MPI-INF-3DHP \cite{mehta2017monocular}. Mention that our method not only significantly improves the accuracy of single-frame 3D pose estimation but also outperforms even the accuracy of 3D pose estimation networks based on temporal information. Our contribution can be summarized as follows:

\begin{itemize}
\item We propose two novel insights about 3DHPE methods involving image information. For one thing, overly focusing on dataset-biased background leads to poor generalization ability. For another, valuable image patches for estimating specific keypoint's 3D coordinates are not confined to its exact image location; they extend to areas with structurally related positions.

\item We propose a 3DHPE framework leveraging effective image features in two stages: broad query followed by focused exploration. It not only enables keypoints to determine all the necessary image features but also prevents excessive training on the background, thus improving generalization.

\item We propose a novel Pose-guided Transformer Layer that effectively improves the keypoints' ability to significant features. Besides, we propose an Adaptive Feature Selection Module, which adaptively stops irrelevant image features from further training.
\end{itemize}

\section{Related work}
In the past few years, there has been extensive research on deep-learning-based algorithms for monocular 3D human pose estimation. Methods that directly regress 3D pose from the image are popular in the early stages \cite{li20153d}. However, these approaches suffered from limited performance due to their reliance on training and testing within the constraints of 3D Motion Capture data \cite{xu2021monocular}. To address this limitation, the ``lifting" method emerged as the dominant approach, offering better solutions to the problem.

\subsection{``Lifting" Based 3D Human Pose Estimation}
``Lifting" based approaches leverage off-the-shelf 2D human pose estimators trained on large and more diverse 2D datasets. By adopting this, the process of 3D human pose estimation is simplified to lifting the 2D pose to 3D pose without image participation. \citeauthor{martinez2017simple} first proposed a fully connected residual network in this approach. To handle the issue of depth ambiguity in the lifting process, some methods have leveraged temporal information \cite{pavllo20193d, chen2021anatomy} or proposed models with multiple hypotheses \cite{li2019generating, li2022mhformer}.
\begin{figure}[t]
  \centering
%   \fbox{\rule{0pt}{2in} \rule{0.95\columnwidth}{0pt}}
  \includegraphics[width=\linewidth]{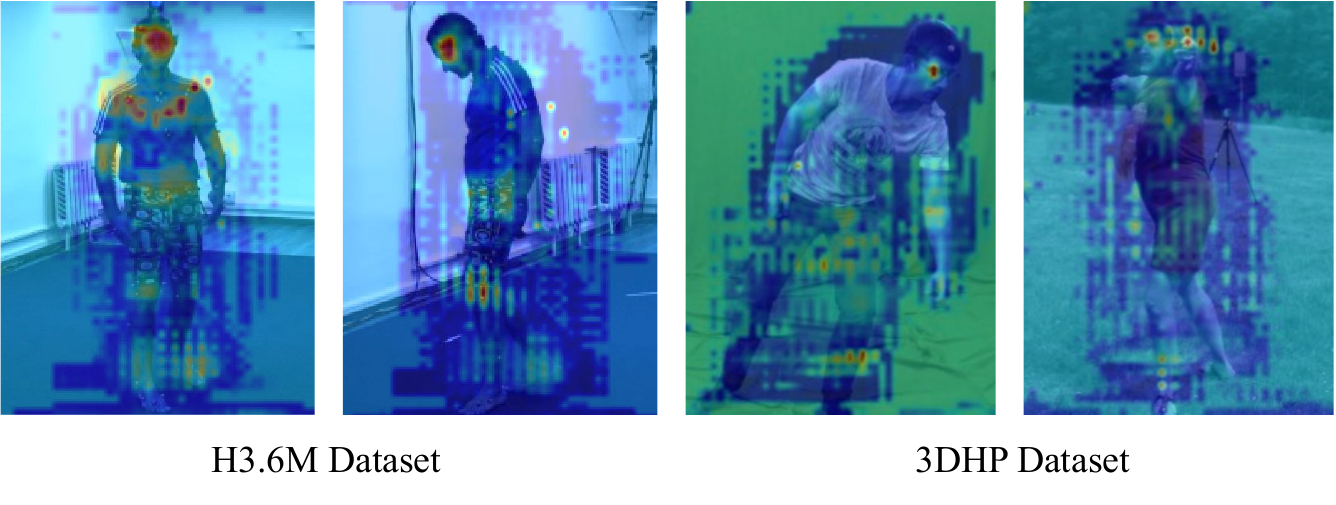}
  \caption{Examples of attention map visualization of all keypoints on image features in two datasets.}
  \label{fig:analysis1}
\end{figure} 
\begin{table}
  \centering
\renewcommand\arraystretch{1.1}
\begin{tabular}{c|ccc}
\hline
Traindata & Testdata & MPJPE$\downarrow$ & Background Attention \\ \hline
H3.6M     & H3.6M    & 30.4  & 73\%                 \\ \hline
H3.6M     & 3DHP     & 74.2  & 75\%        \\ \hline
\end{tabular}
    \caption{
    Comparison of different test datasets and observed issues of excessive attention to the background and poor generalization.}
  \label{table:analysis1}
\end{table}
\begin{figure}[t]
  \centering
%   \fbox{\rule{0pt}{2in} \rule{0.95\columnwidth}{0pt}}
  \includegraphics[width=\linewidth]{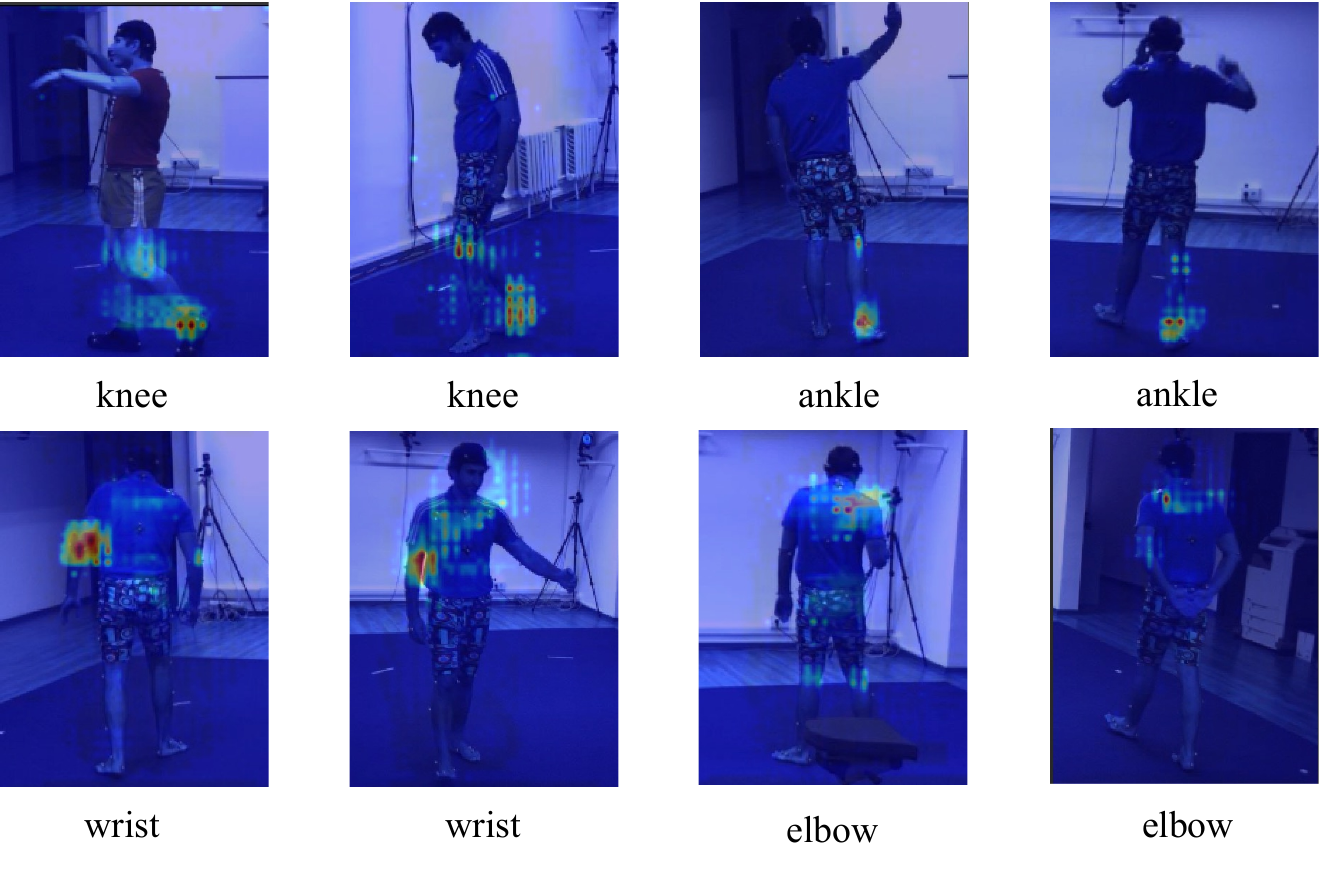}
  \caption{Visualization examples of heatmaps depicting the attention of specific keypoints.}
  \label{fig:analysis2}
\end{figure} 
\subsection{Fusion Approach}
Apart from the two mainstream approaches, there exist some methods that combine 2D pose with image information. Despite the remarkable attempts made by these methods, they still exhibit certain limitations. For example, some methods did not leverage off-the-shelf 2D pose estimators to generate 2D pose \cite{zhao2019semantic, liu2019feature}. These approaches not only add a burden to the network but also fail to leverage the benefits of 2D estimators mentioned before. Besides, some methods employ rudimentary approaches to integrate image information. For instance, \citeauthor{nie2017monocular, xu2021monocular} segment image patches around keypoint positions to assist in generating the 3D pose. Similarly, \citeauthor{zhao2019semantic, liu2019feature} overlay image features extracted from keypoints position onto 2D keypoints. Nevertheless, the insight proposed in the next section proves this local concatenation approach might be ineffective. Moreover, \citeauthor{zhou2019hemlets, gong2023diffpose} utilize 2D keypoints heatmap on image to provide extra information. Indeed, heatmap only contains limited information, which may not be sufficient to accurately regress 3D poses.

\section{Insight of Image effect to 3DHPE}
In this section, we study the roles and limitations of image features in estimating 3D pose using attention mechanisms. The keypoint-to-image attention map represents which image patches offer beneficial information for estimating the 3D coordinates of that keypoint.

\subsection{Background Overfitting}
Given the task of estimating the relative coordinates of human keypoints, the presence of non-contact backgrounds in 3D capture-environment datasets can be considered a form of dataset-biased noise. When we visualized the average attention maps of keypoints on the images, we observed a wide-ranging and indiscriminate focus on background features, as shown in Fig \ref{fig:analysis1}. This reveals the model's overfitting to background information, which could be a potential cause of the poor generalization of image-based models. We further quantified the proportion of attention on background features, and found very high proportions on both datasets (73\%, 75\%), as shown in Table \ref{table:analysis1}. 

\begin{figure*}[t]
  \centering

  \includegraphics[width=\linewidth]{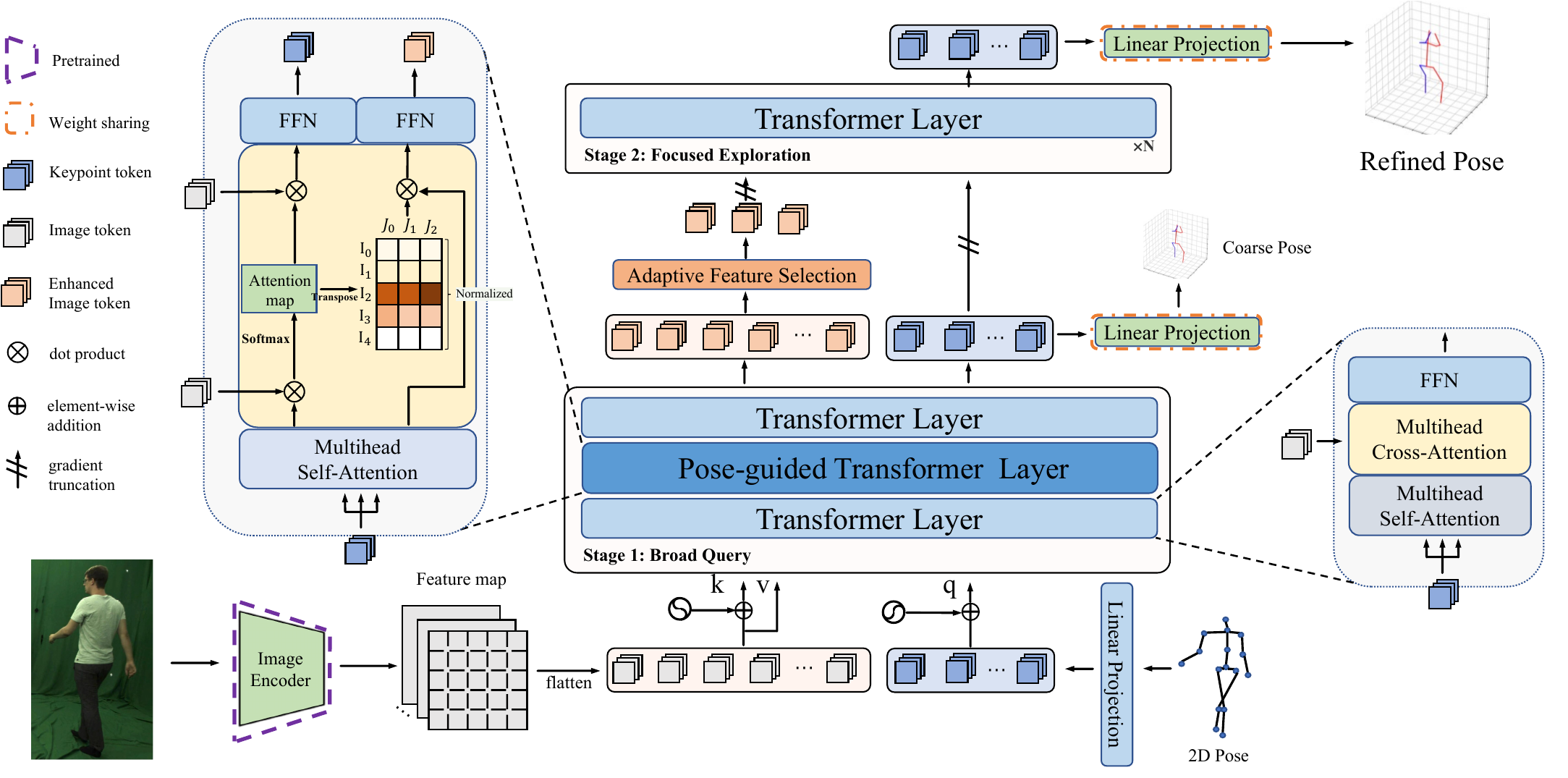}

  \caption{The overview of the proposed network.
  }
  \label{fig:framework}
\end{figure*}

\subsection{Structural Assistance}
Logically speaking, for a specific human body keypoint in the image, the keypoint's own image features can only provide its 2D coordinates in the image. However, the relative depth coordinate with respect to the pelvis point requires prior knowledge derived from combining other human structure features. We present attention of specific keypoints on image features, examples shown in Fig \ref{fig:analysis2}. Not surprisingly, our findings lead to the conclusion: not only the image features of keypoint's locations are required. It will extend to body structure positions that provide depth information for that keypoint. For instance, the \emph{knee} keypoint gives attention to ankles, the \emph{wrist} keypoint gives attention to elbows and shoulders, and the \emph{elbow} keypoint gives attention to shoulders. Hence, previous methods have been mistaken in their assumption that only concatenating image patches or features around keypoints is sufficient.

\section{Method}
In this section, we provide a detailed description of the proposed framework, as illustrated in Fig \ref{fig:framework}. Given a 2D pose $J_{2d} \in \mathbb{R}^{N\times 2}$, our method aims to reconstruct the 3D pose $J_{3d} \in \mathbb{R}^{N\times 3}$ by effectively leveraging the information from a cropped image $I \in \mathbb{R}^{h\times w \times 3}$, where $N$ is the number of keypoints, $h, w$ is the input image size. To accomplish this, we proposed a Progressive Training framework. It consists of two stages. In Stage 1, we allow the keypoints to query beneficial information from all image features under coarse pose supervision until convergence. Subsequently, to counteract the detriment of background features' excessive training on generalization, we employ an Adaptive Feature Selection Module to prune less crucial image features. Then, in Stage 2, keypoints exclusively query the preserved image features to generate a refined pose.

\subsection{Stage 1: Broad Query}
The image $I$ is fed into a 2d-pose-estimation-pretrained image encoder, resulting in features $F_I \in \mathbb{R}^{H\times W \times d}$, which are flattened into tokens $T_I \in \mathbb{R}^{HW \times d}$ with sequence length $HW$ and dimension $d$. Similarly, the 2D pose $J_{2d}$ is transformed to pose tokens $T_P \in \mathbb{R}^{N \times d}$ by linear projection.

Subsequently, the image tokens and keypoint tokens are fed into three consecutive Transformer Layers. Situated in the middle is the specially crafted Pose-guided Transformer Layer, intended to selectively enhance image tokens while diminishing the keypoints' focus on irrelevant image tokens. The resulting keypoint tokens are then projected linearly to generate the coarse 3D pose denoted as $J_{3d_1} \in \mathbb{R}^{J \times 3}$.

\noindent\textbf{Transformer Layer}
 consists of three consecutive modules, including Multi-head Self-Attention (MSA), Multi-head Cross-Attention (MCA), and Feed Forward Network (FFN). Multi-head Attention can be formulated as:
\begin{equation}
    A({Q}, {K}, {V}) = softmax(\frac{{Q}\cdot {K}^\intercal}{\sqrt{d}}) \cdot {V}
\end{equation}
In MSA, keypoints tokens are linearly mapped to Queries ${Q} \in \mathbb{R}^{N\times d}$, Keys ${K}\in \mathbb{R}^{N\times d}$, and Values ${V} \in \mathbb{R}^{N\times d}$. Similarly, in MCA, keypoints tokens are linearly mapped to Queries ${Q} \in \mathbb{R}^{N\times d}$, Image tokens are linearly mapped into Keys ${K}\in \mathbb{R}^{HW\times d}$, and Values ${V} \in \mathbb{R}^{HW\times d}$.

\noindent\textbf{Pose-guided Transformer Layer.}
We designed a Pose-guided dual attention structure that effectively reduces the keypoints' attention to the background. It leverages the pose-to-image attention matrix to allow the image features to reversely query and aggregate the keypoints features. By our design, the more crucial image patches can obtain more information from the keypoint features. 

Specifically, the novel attention mechanism produces two outputs: keypoint tokens $\hat{T_J}$ and enhanced image tokens $\hat{T_I}$. The update of image tokens will be influenced by the update of keypoint tokens through Attention Map ($\mathbb{A}$). The formulations are as follows:
\begin{equation}
\begin{aligned}
    \mathbb{A} = softmax(\frac{{Q}\cdot {K}^\intercal}{\sqrt{d}}) \\
    \hat{T_J} = \mathbb{A} \cdot {V}_I + T_J \\
    \hat{T_I} = \mathbb{A}^\intercal \cdot {V}_J + T_I
    \end{aligned}
\end{equation}
Similarly, keypoints tokens are linearly mapped to Queries ${Q} \in \mathbb{R}^{N\times d}$, Image tokens are linearly mapped into Keys ${K}\in \mathbb{R}^{HW\times d}$. ${V}_I \in \mathbb{R}^{HW\times d}$ and ${V}_J\in \mathbb{R}^{N\times d}$ are Values linearly mapped from images tokens and keypoints tokens, respectively. The attention map $\mathbb{A} \in \mathbb{R}^{N \times HW}$  represents the weighting that keypoint tokens assign to image tokens, and it is normalized using the $softmax$ function. Similarly, the transposed attention map $\mathbb{A}^\intercal \in \mathbb{R}^{HW \times N}$ represents the weights image tokens assign to keypoint tokens. Because of the normalization before transposition, intuitively, the image tokens deemed more important receive a greater overall weight from the keypoint tokens. This signifies that image tokens with higher significance can gather more information from keypoint tokens, while on the contrary, less significant image tokens (typically background tokens) collect limited information.
\begin{figure}[t]
  \centering
%   \fbox{\rule{0pt}{2in} \rule{0.95\columnwidth}{0pt}}
  \includegraphics[width=\linewidth]{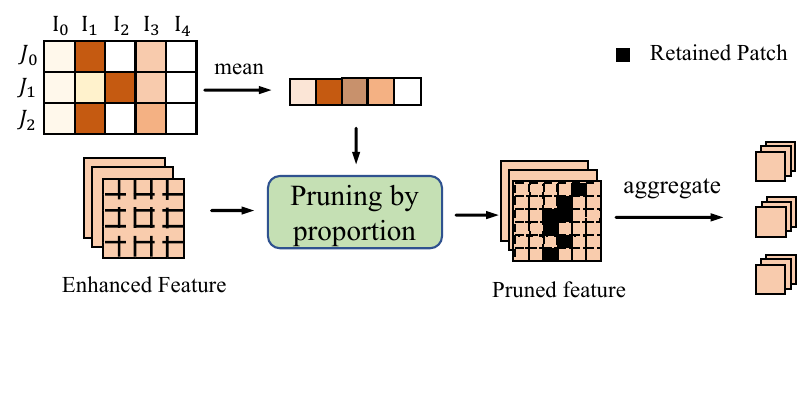}
  \caption{Details of Adaptive Feature Selection Module.}
  \label{fig:KeyFeature}
\end{figure}

We choose to replace only the second Transformer Layer with the proposed Pose-guided Transformer Layer for the following reasons: before guiding the update of image features, the keypoint tokens require an initial perception of the image features (by the first transformer layer) to evaluate their significance. The final layer cannot be replaced, as the update of image tokens lacks direct supervision in Stage 1.
\subsection{Adaptive Feature Selection Module}
To prevent less significant image tokens from further training and aggregating critical image tokens. We propose an Adaptive Feature Selection Module, details shown in Fig. \ref{fig:KeyFeature}. We leverage the attention map from the last transformer layer to rank the importance of image tokens. For simplicity, we aggregate the attention weights of all keypoints on image features and set a retention rate, denoted as $r$ ( $0 < r < 1$). The top $r \times HW$ image tokens with the highest weights are retained.
\begin{table*}[htb]
  \normalsize
  \centering
  \resizebox{\textwidth}{!}{
  \begin{tabular}{@{}l|ccccccccccccccc|c@{}}
  \hline
  Method & Dir. & Disc & Eat & Greet & Phone & Photo & Pose & Purch. & Sit & SitD. & Smoke & Wait & WalkD. & Walk & WalkT. & Avg.\\
  \hline\hline
  
Learning \cite{fang2018learning} &50.1 &54.3 &57.0 &57.1 &66.6 &73.3 &53.4 &55.7 &72.8 &88.6 &60.3 &57.7 &62.7 &47.5 &50.6 &60.4\\

  % Semantic Graph Convolutional Networks for 3D Human Pose Regression
SemGCN \cite{zhao2019semantic}$*$ &\underline{47.3}	&60.7	&\underline{51.4}	&60.5	&\underline{61.1}	&\underline{49.9}	&\textbf{47.3}	&68.1	&86.2	&\underline{55.0}	&67.8	&61.0	&\underline{42.1}	&60.6	&\underline{45.3}	&\underline{57.6} \\

% Monocular 3D Pose Estimation via Pose Grammar and Data Augmentation
Monocular \cite{xu2021monocular}$*$ &\textbf{47.1}	&\underline{52.8}	&54.2	&\underline{54.9}	&63.8	&72.5	&51.7	&\underline{54.3}	&\underline{70.9}	&85.0	&58.7	&\underline{54.9}	&59.7	&\textbf{43.8}	&47.1	&58.1	\\ 

% Graformer: Graph convolution transformer for 3d pose estimation
Graformer \cite{zhao2022graformer}	&49.3	&53.9	&54.1	&55.0	&63.0	&69.8	&51.1	&\textbf{53.3}	&\textbf{69.4}	&90.0	&\underline{58.0}	&55.2	&60.3	&\underline{47.4}  &50.6	&58.7 \\
\hline
Ours $*$ 	&48.3	&\textbf{51.5}	&\textbf{46.1}	&\textbf{48.5}	&\textbf{53.7}	&\textbf{42.8}	&\textbf{47.3}	&59.9	&71.0	&\textbf{51.6}	&\textbf{52.7}	&\textbf{46.1}	&\textbf{39.8}	&53.0	&\textbf{43.9}	&\textbf{51.0} \\
\hline\hline

VideoPose \cite{pavllo20193d} $\dagger$ &45.1	&47.4	&\underline{42.0}	&46.0	&49.1	&56.7	&44.5	&\underline{44.4}	&57.2	&66.1	&\underline{47.5}	&44.8	&49.2	&\underline{32.6}	&\underline{34.0}	&47.1\\ 

% Graph Stacked Hourglass Networks for 3D Human Pose Estimation
GraphSH \cite{xu2021graph} &45.2 &49.9 &47.5 &50.9 &54.9 &66.1 &48.5 &46.3 &59.7 &71.5 &51.4 &48.6 &53.9 &39.9 &44.1 &51.9 \\

% Modulated Graph Convolutional Network for 3D Human Pose Estimation6
MGCN \cite{zou2021modulated} &45.4 &49.2 &45.7 &49.4 &50.4 &58.2 &47.9 &46.0 &57.5 &63.0 &49.7 &46.6 &52.2 &38.9 &40.8 &49.4 \\

MHFormer \cite{li2022mhformer}$\dagger$ f=243 &\textbf{39.2}	&\textbf{43.1}	&\textbf{40.1}	&\textbf{40.9}	&\textbf{44.9}	&\underline{51.2}	&\textbf{40.6}	&\textbf{41.3}	&\textbf{53.5}	&\underline{60.3}	&\textbf{43.7}	&\textbf{41.1}	&\underline{43.8}	&\textbf{29.8}	&\textbf{30.6}	&\textbf{43.0} \\

Pose-Oriented \cite{li2023pose} &47.9 &50.0 &47.1 &51.3 &51.2 &59.5 &48.7 &46.9 &\underline{56.0} &61.9 &51.1 &48.9 &54.3 &40.0 &42.9 &50.5\\

diffPose \cite{gong2023diffpose}	&\underline{42.8}	&49.1	&45.2	&48.7	&52.1	&63.5	&46.3	&45.2	&58.6	&66.3	&50.4	&47.6	&52.0	&37.6	&40.2	&49.7\\
  \hline
Ours$*$	&44.9	&\underline{46.4}	&42.4	&\underline{44.9}	&\underline{48.7}	&\textbf{40.1}	&\underline{44.3}	&55.0	&58.9	&\textbf{47.1}	&48.2 &\underline{42.6}	&\textbf{36.9}	&48.8	&40.1	&\underline{46.4}\\

\hline
\hline
SemGCN \cite{zhao2019semantic}$*$ &37.8	&49.4	&37.6	&40.9	&45.1	&41.4	&40.1	&48.3	&50.1	&42.2	&53.5	&44.3	&40.5	&47.3	&39.0	&43.8 \\

VideoPose \cite{pavllo20193d} $\dagger$   &-&-&-&-&-&-&-&-&-&-&-&-&-&-&- &37.2\\

GraphSH \cite{xu2021graph} &35.8	&38.1	&31.0	&35.3	&35.8	&43.2	&37.3	&31.7	&38.4	&45.5	&35.4	&36.7	&36.8	&27.9	&30.7	&35.8 \\
Graformer \cite{zhao2022graformer} &32.0	&38.0	&30.4	 &34.4	&34.7	&43.3	&35.2	&31.4	&38.0	&46.2	&34.2	&35.7	&36.1	&27.4	&30.6	&35.2 \\
MHFormer \cite{li2022mhformer} $\dagger$ f=243&\textbf{27.7}	&\underline{32.1}	&29.1	&\textbf{28.9}	&\underline{30.0}	&\underline{33.9}	&33.0	&31.2	&37.0	&\underline{39.3}	&\textbf{30.0}	&\underline{31.0}	&\underline{29.4}	&\textbf{22.2}	&\textbf{23.0}	&\underline{30.5}\\
Pose-Oriented \cite{li2023pose} &32.9 &38.3 &28.3 &33.8 &34.9 &38.7 &37.2 &30.7 &34.5 &39.7 &33.9 &34.7 &34.3 &26.1 &28.9 &33.8\\
diffPose \cite{gong2023diffpose} &\underline{28.8}	&32.7	&\underline{27.8}	&30.9	&32.8	&38.9	&\underline{32.2}	&\textbf{28.3}	&\underline{33.3}	&41.0	&\underline{31.0}	&32.1	&31.5	&25.9	&27.5	&31.6 \\
\hline
 Ours$*$ &29.5	&\textbf{30.1}	&\textbf{25.0}	&\underline{29.0}	&\textbf{28.5}	&\textbf{28.6}	&\textbf{26.9}	&\underline{30.5}	&\textbf{31.1}	&\textbf{27.7}	&32.4	&\textbf{27.7}	&\textbf{24.8}	&\underline{30.0} 	&\underline{25.9} 	&\textbf{28.6} \\
 
  \hline
  \end{tabular}
}
\caption
  {
    Quantitative comparison with the state-of-the-art methods on Human3.6M under Protocol 1, using SH (Stacked Hourglass network \cite{newell2016stacked}) detected 2D poses (top), using CPN (cascaded pyramid network \cite{chen2018cascaded}) detected 2D poses (middle), ground truth 2D poses (bottom) as inputs. 
    ($\dagger$) - uses temporal information.  $(*)$ - uses image information.
    \textbf{Blod}: best; 
    \underline{Underlined}: second best.  
  } 
  \label{table:h36m}

\end{table*}

\begin{table}
  \centering

\renewcommand\arraystretch{1.1}
  \begin{tabular}{@{}l|ccc@{}}
  \hline
  Method & PCK $\uparrow$ & AUC $\uparrow$ & MPJPE $\downarrow$ \\
  \hline\hline
Simple \cite{martinez2017simple} &82.6 &50.2 &88.6 \\ 
  Cascaded \cite{li2020cascaded}&81.2 &46.1 &99.7 \\
  MGCN \cite{zou2021modulated} &86.1 &53.7 &- \\
  Pose-Oriented \cite{li2023pose} &84.1 &53.7 &- \\

  \hline
  
  Ours &\textbf{88.2} &\textbf{59.3} &\textbf{68.9} \\
  \hline
  
  \end{tabular}
\caption{
    Quantitative comparison with the state-of-the-art methods on MPI-INF-3DHP. 
    Best in \textbf{Bold}.
  }
  \label{table:3dhp}
  
\end{table}

\subsection{Stage 2: Focused Exploration}
In this stage, we allow the keypoint tokens to further dig information from selected critical image tokens and generate a refined 3D pose $J_{3d_2} \in \mathbb{R}^{J \times 3}$.  Specifically, we freeze the weights trained in the former stage and feed the keypoints tokens and selected image tokens into a new Transformer Block consisting of several Transformer Layers. Then, the output keypoint tokens will be projected to a refined 3D pose $J_{3d_2}$by Linear Projection.
\subsection{Loss Function}
Our model is traind with Mean Squared Error (MSE) loss.
\begin{equation}
    {L} = \sum^{J}_{i=1} \lVert Y_i - \hat{Y_i} \rVert_2
\end{equation}

where $Y_i$ and $\hat{Y_i}$ represent the predicted and ground-truth 3D pose of joint $i$, respectively.

\section{Experiments}
\subsection{Datasets and Evaluation Metrics}
We evaluate our method on two widely-used datasets for 3DHPE: Human3.6M \cite{ionescu2013human3} and MPI-INF-3DHP \cite{mehta2017monocular}.

\noindent\textbf{Human3.6M (H3.6M)} is the largest and most representative benchmark for 3DHPE. Following \citeauthor{martinez2017simple}, we use subject S1, S5, S6, S7 and S8 for training, and S9, S11 for testing. We down-sampled the original frame rate from 50 fps to 5 fps for faster training. The Mean Per Joint Position Error (MPJPE) is computed under two protocols: Protocol 1 computes MPJPE between ground truth and the estimated 3D poses after aligning their root (pelvis) keypoints; Protocol 2 is the MPJPE after aligning the estimated 3D pose with the ground truth using translation, rotation, and scale (P-MPJPE). 

\noindent\textbf{MPI-INF-3DHP (3DHP)} provides monocular videos of six subjects acting in three different scenes, including indoors and outdoors. This dataset is often used to evaluate the generalization performance of different models. Following the convention, we directly apply our model trained on H36M dataset to this dataset without re-training. We report results using three metrics, Mean Per Joint Position Error (MPJPE), Percentage of Correctly estimated Keypoints (PCK) with a threshold of 150 mm, and Area Under the Curve (AUC) a range of PCK thresholds.

\subsection{Implementation Details}
We take HRNet-w32 as our backbone with input size $256 \times 192$, which is pretrained on MS COCO 2017 dataset \cite{lin2014microsoft}, provided by \citeauthor{sun2019deep}. The retention $r$ is set to 0.3. The number of Transformer layers in Stage 2 is set to 3. For a fair comparison, following previous work \cite{pavllo20193d, martinez2017simple}, we obtain 2D pose detections cascaded pyramid network (CPN) \cite{chen2018cascaded} and stacked hourglass network (SH) \cite{newell2016stacked}. We take the ground-truth bounding boxes. Our model is implemented in Pytorch and optimized via Adam. All experiments are conducted on two NVIDIA RTX 3090 GPUs. The initial learning rate is set to 0.001 with a shrink factor of 0.9 per 4 epochs with 128 batch size. We first train the initial interaction stage and image encoder for 20 epochs and freeze them to train the remaining modules.

\subsection{Comparison with the State-of-the-art Methods}
\noindent\textbf{Results on Human3.6M.}  The proposed method is compared with the state-of-the-art methods on Human3.6M. The result and comparison of our model with SH detected, CPN detected and ground-truth 2D pose are reported in Table \ref{table:h36m}. Without bells and whistles, our method outperforms all previous state-of-the-art methods by a large margin under protocol 1 for all single-frame methods. Besides, our method even achieves comparable results with video-based method. 

\noindent\textbf{Result on MPI-INF-3DHP.} To assess the generalization ability, we directly applied our method on MPI-INF-3DHP datasets trained on Human3.6M dataset without fine-tuning. We take 2D ground-truth input. Table \ref{table:3dhp} shows the result of different methods. Our approach achieves the best performance on all metrics (PCK, AUC, and MPJPE). It emphasizes the strong generalization ability of our model.

\begin{figure*}[t]
  \centering
  \includegraphics[width=\linewidth]{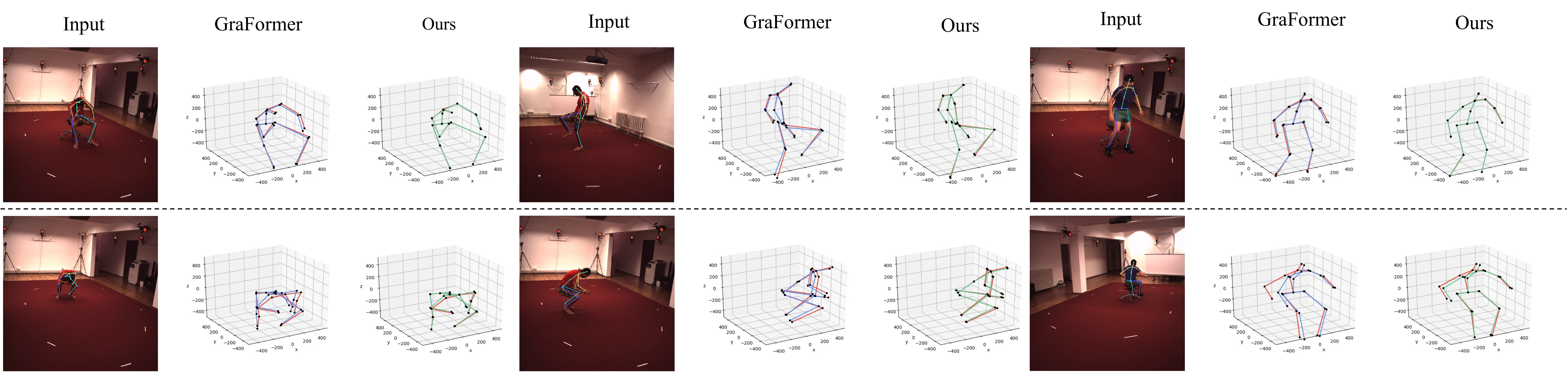}
  \caption{Qualitative results on Human3.6M. Green lines represent our results, results from baseline (Graformer) are represented by blue lines, and Ground-truth is represented in red. Easy case examples are shown at the top, and hard case examples are shown at the bottom.
  }
  \label{fig:Qualitative}
\end{figure*}

\noindent\textbf{Qualitative Result}.
Figure \ref{fig:Qualitative} demonstrates some qualitative results on Human3.6M dataset compared with Graformer  \cite{zhao2022graformer} with 2d ground-truth input. Our result can achieve almost the same as the ground-truth when faced with easy cases. For hard cases, compared with the baseline, our result can effectively reduce the gap between the estimated result and ground-truth. 

\subsection{Ablation study and Discussion}

\noindent\textbf{Effect on Progressive Learning strategy.} We first compared the Coarse Pose and Refined Pose results on both H3.6M and 3DHP on Table \ref{table:ablation_refine}. It can be seen that Stage 2 brings 0.6mm performance on Human3.6m and 2.0mm performance on 3DHP. To demonstrate that performance improvement is not brought by the increased model parameters, we retrained our method from end to end without coarse pose supervision or with both supervisions. The performance drops 4.0mm or 1.6mm in Human3.6m and drops 7.5mm or 3.5mm in 3DHP.

\begin{figure}[t]
  \centering
%   \fbox{\rule{0pt}{2in} \rule{0.95\columnwidth}{0pt}}
  \includegraphics[width=\linewidth]{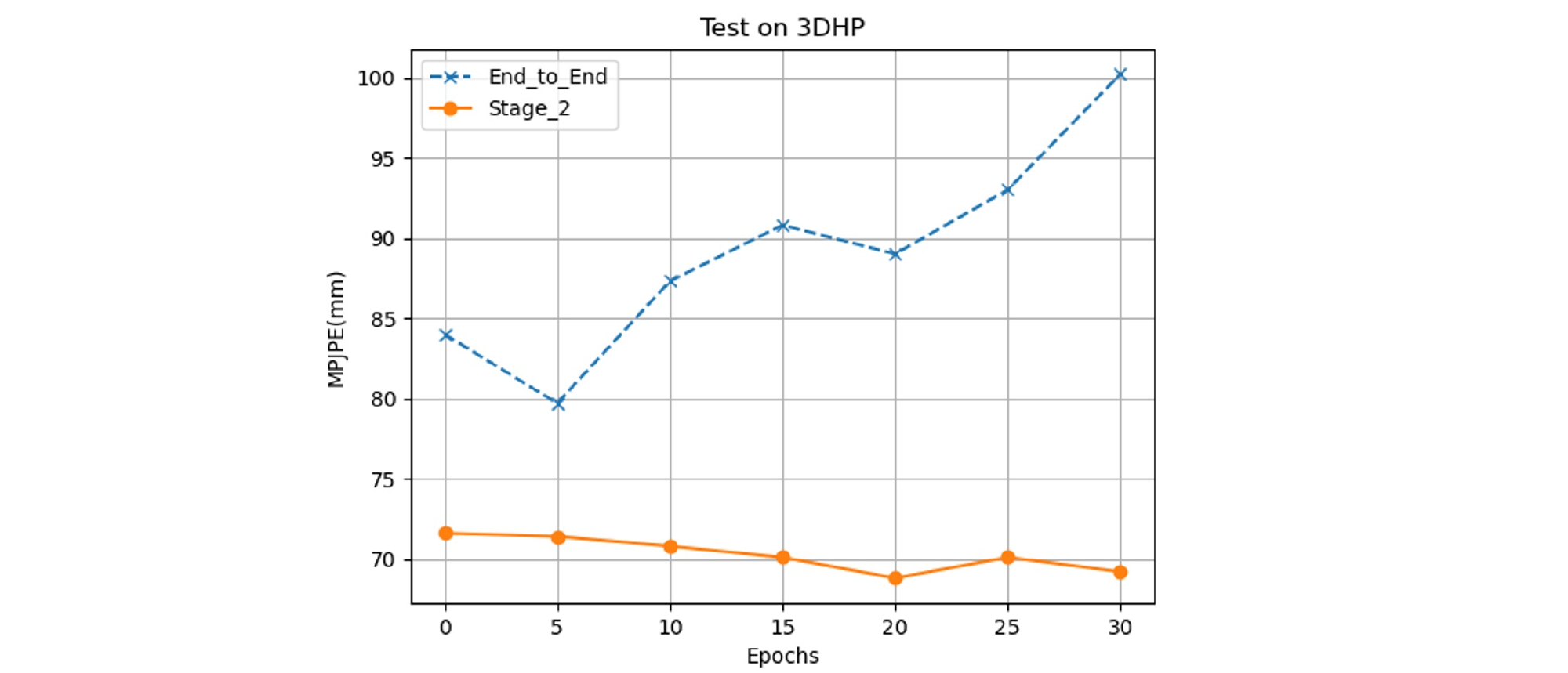}
  \caption{Effect of progressive learning strategy on test curves on 3DHP dataset.}
  \label{fig:ablation_curve}
\end{figure}

To demonstrate the ability of our method to improve generalization, we compare the test result curves of end-to-end training and progressive training on the 3DHP dataset as shown in Fig. \ref{fig:ablation_curve}. It can be seen that in the case of training only on key features (Stage 2), the test accuracy of the 3DHP dataset continues to increase (lower in MPJPE). This demonstrates that preventing excessive training on background features can enhance generalization.
\begin{table}
  \centering
\renewcommand\arraystretch{1.1}
  \begin{tabular}{c|cc}
    \hline
    Strategy &  Human3.6m & 3DHP\\
        \hline
        end-to-end w/ fine supervision & 32.6& 76.4\\
        end-to-end w/ both supervision & 30.2&72.4\\
     Coarse Pose & 29.2 &70.9 \\
    Refined Pose & \textbf{28.6} &\textbf{68.9}\\
  \hline
  \end{tabular}
    \caption{
    Ablation study on progressive learning.}
  \label{table:ablation_refine}
\end{table}

\begin{table}
  \centering
\renewcommand\arraystretch{1.1}
\begin{tabular}{c|ccc}
   \hline
& dataset & MPJPE & Background Attention \\    \hline
w/o PGTL & H3.6M     & 30.4    & 73\%                    \\ 
& 3DHP   & 74.2        & 75\%      \\    \hline
w/ PGTL  & H3.6M   & \textbf{29.2}  &\textbf{60\%}\\ 
& 3DHP      & \textbf{70.9}                       & \textbf{63\%}                               \\      \hline
\end{tabular}
    \caption{
    Effect on the Pose-guided Transformer Layer (PGTL). Represented in Coarse pose in MPJPE.}
  \label{table:ablation1}
\end{table}
\begin{table}
  \centering
\renewcommand\arraystretch{1.1}
  \begin{tabular}{ccc}
   \hline
    Transformer Layer & Pose-guided Layer & MPJPE\\
   \hline
        0 & 2& 30.4\\
     1 & 1 &\textbf{29.2} \\
    2 & 0 & 31.3 \\
   \hline
  \end{tabular}
    \caption{
    Effect on number of Transformer Layers in Stage 1.}
  \label{table:ablation_layers}
\end{table}
\begin{figure}[t]
  \centering
%   \fbox{\rule{0pt}{2in} \rule{0.95\columnwidth}{0pt}}
  \includegraphics[width=\linewidth]{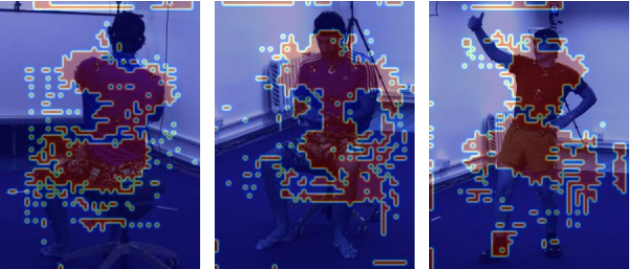}
  \caption{Visualization of the retained feature (red pixels).}
  \label{fig:ablation_r}
\end{figure}

\noindent\textbf{Ablation study on Pose-guide Transformer Layer.}  We then diagnose the effect of the designed Pose-guide Transformer Layer. To test the ability to reduce the focus of keypoints on background information, we conducted a quantitative analysis. We took a 30-pixel-radius circle for the positions of each keypoints on the attention map, and defined these areas outside these circles as ``background". We then calculated the average attention of the keypoints to the background and performed statistical analysis on the test set. The result is shown in Table \ref{table:ablation1}. We achieved 1.2mm on H3.6m improvement and 3.3mm improvement on 3DHP by replacing Tansformer Layer with Pose-guide Transformer Layer. More importantly, the attention rate to background in the first stage dropped from 73\% to 60\% on H3.6m dataset, and from 75\% to 63\% on 3DHP dataset, which shows our approach is effective in enhancing the ability of keypoints to perceive critical features. In addition, we performed an ablation study on the number of layers in the Pose-guided Transformer Layer, shown in Table \ref{table:ablation_layers}. 

\noindent\textbf{Effect on Adaptive Feature Selection Module.} We then conduct an ablation study on Adaptive Feature Selection Module. We first tested the effect of different retention rates ($r$) on the results, shown in Table \ref{table:ablation_r}.

It can be seen that we reach the best result when $r$ is set to $0.3$. Besides, it is worth noting that we achieve 0.3mm and 0.2mm improvement when $r$ is set to 0.01 and 1. The former can be explained by the self-attention in Stage 2, which further models the dependency between keypoints. The latter demonstrates the effect of the selection operation. We further visualize some examples of pruning on feature maps when $r$ is set to 0.3, shown in Fig \ref{fig:ablation_r}. It shows the module effectively prunes the background feature.
\begin{table}
  \centering
\renewcommand\arraystretch{1.1}
\begin{tabular}{c|cccc}
   \hline
r     & 0.01 & 0.3  & 1    & Coarse Pose \\    \hline
MPJPE & 28.9 & \textbf{28.6} & 29.0 & 29.2        \\    \hline
\end{tabular}
    \caption{
    Ablation study on the retention rate $r$.}
  \label{table:ablation_r}
\end{table}
\section{Conclusion}
This paper gives a new insight into the cause of poor generalization problems and the effectiveness of image features. Based on that, we propose an advanced 3DHPE framework that leverages effective image cues and improves the generalization ability. It comprises two stages: the first involves a broad query for valuable image features, and the second stage focuses on critical features. To accomplish this, we proposed a novel Pose-guided Transformer Layer to reduce the keypoints' attention to background and an Adaptive Feature Selection Module to prune less significant image features. Extensive experiments show that our method achieves state-of-the-art performance on two widely used benchmark datasets and shows great generalization ability. We hope our exploration can provide insights for future 3DHPE research.

\section{Acknowledgement}
This work was supported partly by the National Natural Science Foundation of China (Grant No. 62173045), and the Natural Science Foundation of Hainan Province (Grant No. 622RC675).
\bibliography{aaai24}
\clearpage
\appendix
\section{Appendix}\label{appendix}

\noindent The supplementary is organized as follows. (1) Support materials of the ``Insight" chapter. (2) Additional qualitative and quantitative experiments on Human3.6M and MPI-INF-3DHP. (3) Additional ablation studies.
\subsection{Supplementary Materials for Chapter ``Insight of Image effect to 3DHPE"}

\begin{figure}[t]
  \centering
%   \fbox{\rule{0pt}{2in} \rule{0.95\columnwidth}{0pt}}
  \includegraphics[width=\linewidth]{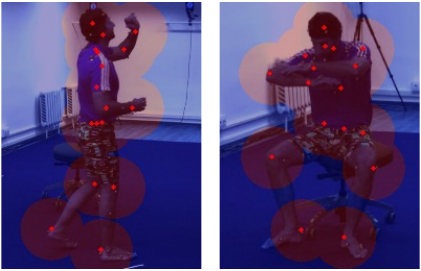}
  \caption{Examples of Background Definition Visualization.}
  \label{fig:appendix1}
\end{figure}

\begin{figure}[t]
  \centering
%   \fbox{\rule{0pt}{2in} \rule{0.95\columnwidth}{0pt}}
  \includegraphics[width=\linewidth]{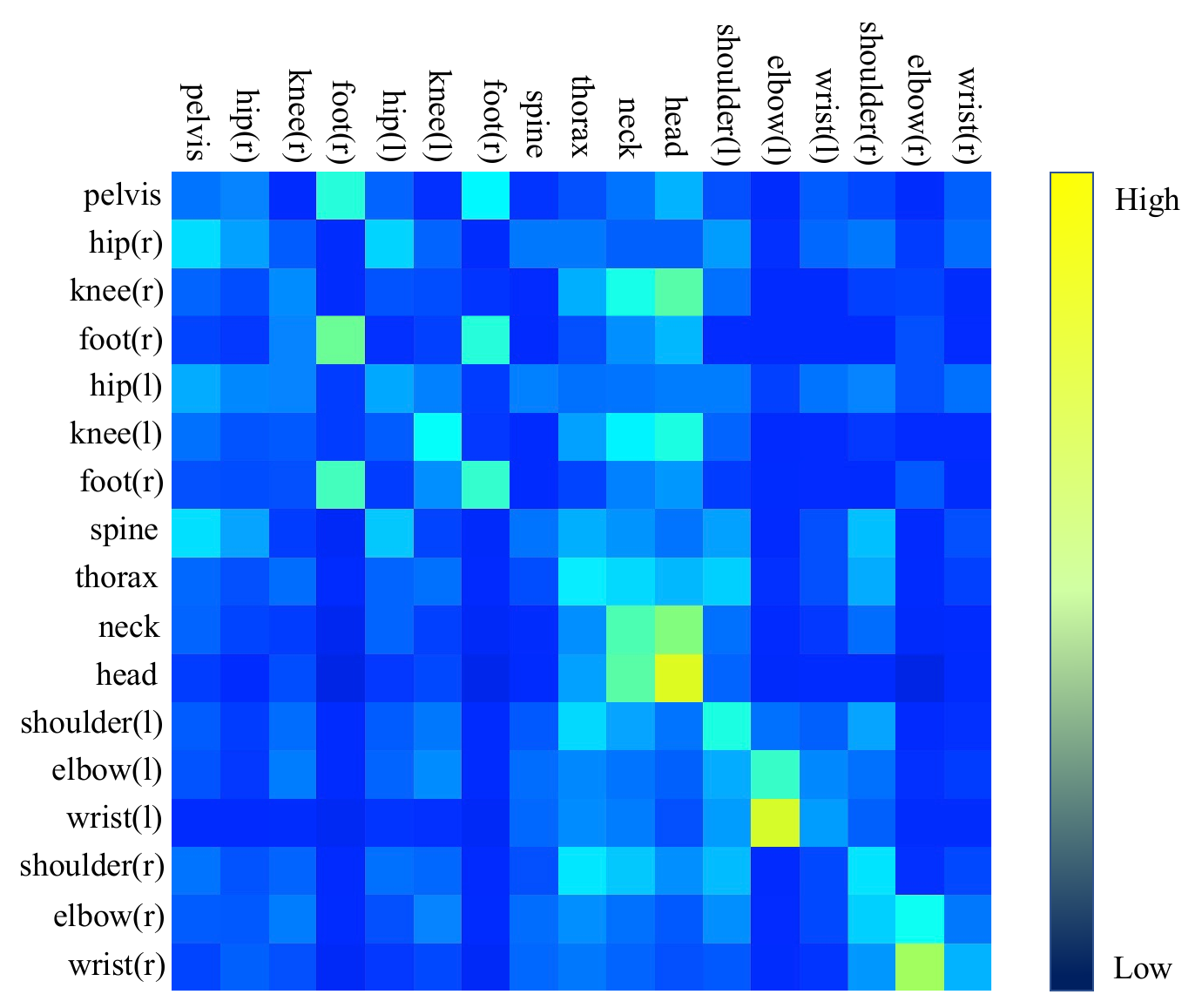}
  \caption{The Attention levels from one keypoint to other keypoints' areas (represented by each row).}
  \label{fig:appendix2}
\end{figure}

\noindent\textbf{Details of the model to generate attention maps.} We straightforwardly stack 3 Transformer Layers. We the attention map in the last Transformer Layer to analyze. For multi-head results, we take the average of all heads.
\vskip 0.05in
\noindent\textbf{Definition of ``Background" in Quantitative Analysis.} We take a 30-pixel-radius circle for the positions of each keypoints on the attention map and define these areas outside these circles as ``background", examples shown in Fig \ref{fig:appendix1}. 
\vskip 0.05in
\noindent\textbf{Supplementary for Structural Assistance} To investigate the role of structural assistance, we quantitatively examined the attention levels of keypoint towards body positions other than their own keypoint areas. By utilizing the circles mentioned above, we divide the human body into $N$ areas (partly overlapped), where $N$ is the number of keypoints. Attentions that fall into these keypoint circles are considered to be the focus on that keypoint area. We conducted a quantitative analysis of all keypoints' attention on the entire H3.6m test set and visualized the results, as shown in Fig \ref{fig:appendix2}. It can be seen that each keypoint exhibits varying degrees of attention toward the areas of other keypoints. It is noticeable that keypoints with low structural correlation, such as \emph{head} and \emph{neck}, tend to concentrate their attention on their respective locations. For highly active keypoints, they exhibit attention towards the image regions of other keypoints, like the \emph{foot(l)} pays attention to the position of \emph{foot(r)} or the \emph{wrist} pays attention to the position of \emph{elbow} in image.

\subsection{Additional Experiments}
\begin{figure*}[t]
  \centering
  \includegraphics[width=\linewidth]{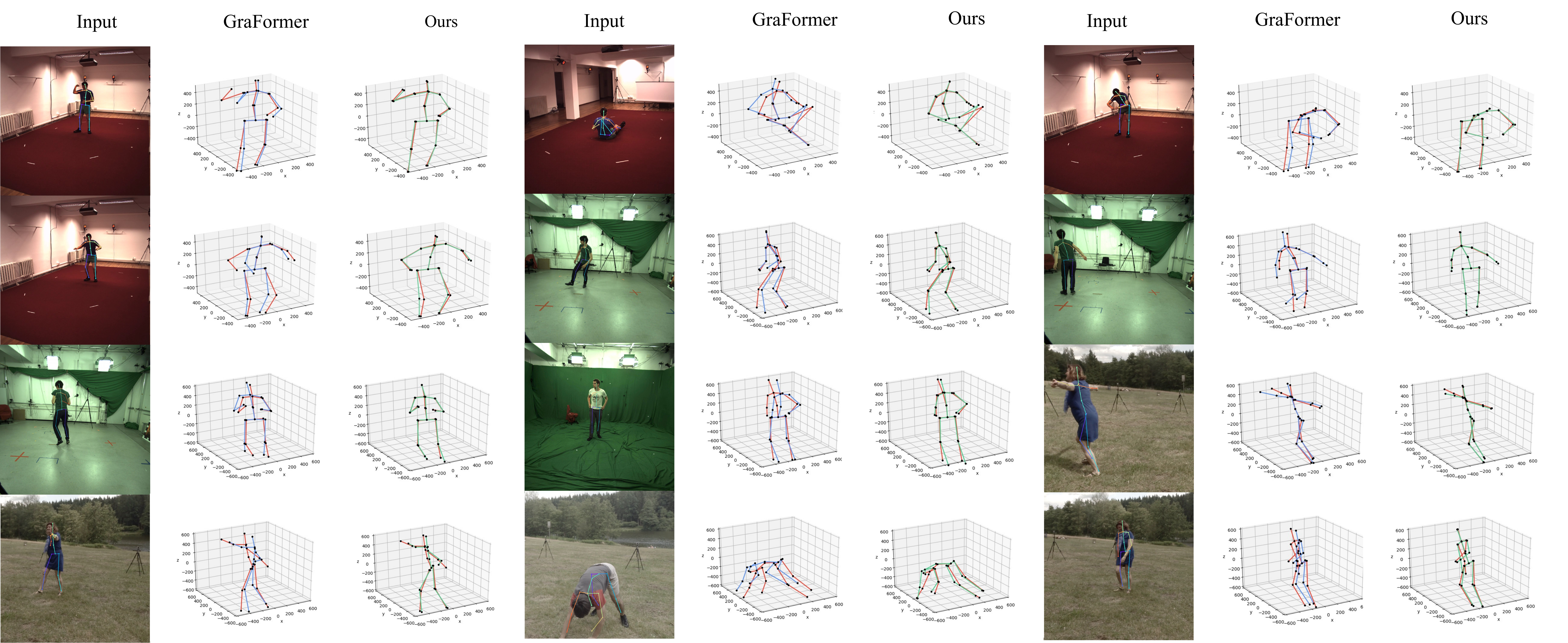}
  \caption{Additional Qualitative results on Human3.6M and MPI-INF-3DHP. Green lines represent our results, results from baseline (Graformer) are represented by blue lines, and the ground-truth is represented in red. The first four are results in H3.6M, the four two are results in 3DHP (in-door), and the last four are results in 3DHP (out-door).
  }
  \label{fig:appendix3}
  
\end{figure*}

\noindent\textbf{Experiment Results on Human3.6M under P-MPJPE (Protocol 2)}. Shown in Table \ref{appendix:pmpjpe}. We present the result taking SH (Stacked Hourglass network) and CPN (Cascaded pyramid network) detected 2D poses as input. Our approach can significantly outperform the state-of-the-art frame-based methods by a large margin. Our results can even be compared to advanced video-based methods (MHFormer), with a difference of just 1.6mm in MPJPE (36.0mm to 34.4mm).
\vskip 0.05 in
\noindent\textbf{Additional qualitative result on Human3.6M and 3DHP}. We additionally qualitatively compared our approach to baseline (Graformer) on both Human3.6M and MPI-INF-3DHP. Show in Fig \ref{fig:appendix3}. Evidently, compared to the baseline, our approach performs better both in indoor and outdoor scenarios, yielding more credible results. The strong performance in outdoor scenarios demonstrates the robust generalization capability of our method. 

\subsection{Additional Ablation Studies}
\noindent\textbf{Effect on Image Encoder Pretraining}
We first conduct an ablation study pertaining to the Image Encoder. We replaced the pretraining of HRNet-w32 with ImageNet pretraining, as shown in Table \ref{table:appendix1}. This resulted in a significant loss of accuracy (from 28.6mm to 35.2mm in MPJPE). This underscores the importance of pretraining the Image Encoder on the 2D pose estimation task, significantly enhancing its perceptual capabilities toward the human body.

\vskip 0.05 in
\noindent\textbf{Ablation study on Image Encoder}
We then conduct experiments with different types of Image Encoders with 2D ground-truth input, as shown in Table \ref{table:appendix2}. All the backbones are pretrained on MS COCO 2017 with 2D human pose estimation task. It is expected that HRNet outperforms ResNet due to its superior capability to retain high-resolution information. Besides, within the same backbone, larger models lead to lower accuracy. It could be hypothesized that stronger models contribute to faster overfitting, thus leading to a reduction in accuracy. 

\begin{table*}[htb]
  \normalsize
  \centering
  \resizebox{\textwidth}{!}{
  \begin{tabular}{@{}l|ccccccccccccccc|c@{}}
  \hline
  Method & Dir. & Disc & Eat & Greet & Phone & Photo & Pose & Purch. & Sit & SitD. & Smoke & Wait & WalkD. & Walk & WalkT. & Avg.\\
  \hline\hline
  
Learning \cite{fang2018learning} &38.2 &41.7 &43.7 &44.9 &48.5 &55.3 &40.2 &\textbf{38.2} &\textbf{54.5} &64.4 &47.2 &44.3 &47.3 &\textbf{36.7} &41.7 &45.7\\
SemGCN \cite{zhao2019semantic}$*$  &-&-&-&-&-&-&-&-&-&-&-&-&-&-&- &-\\
Monocular \cite{xu2021monocular}$*$ &-&-&-&-&-&-&-&-&-&-&-&-&-&-&- &-\\
Graformer\cite{zhao2022graformer}	&-&-&-&-&-&-&-&-&-&-&-&-&-&-&- &-\\
% Graformer: Graph convolution transformer for 3d pose estimation
\hline
Ours $*$	&\textbf{37.5}	&\textbf{38.8}	&\textbf{38.5}	&\textbf{37.8}	&\textbf{42.8}	&\textbf{32.5}	&\textbf{35.6}	&47.7	&56.4	&\textbf{42.2}	&\textbf{40.4}	&\textbf{35.1}	&\textbf{31.2}	&40.9	&\textbf{35.4}	&\textbf{40.1} \\
\hline\hline

VideoPose \cite{pavllo20193d} $\dagger$ &34.2 &36.8 &\underline{33.9} &37.5 &37.1 &43.2 &34.4 &\underline{33.5} &45.3 &52.7 &37.7 &34.1 &38.0 &\underline{25.8} &\underline{27.7} &36.8\\ 
GraphSH \cite{xu2021graph} &-&-&-&-&-&-&-&-&-&-&-&-&-&-&- &-\\
% Modulated Graph Convolutional Network for 3D Human Pose Estimation6
MGCN \cite{zou2021modulated} &35.7 &38.6 &36.3 &40.5 &39.2 &44.5 &37.0 &35.4 &46.4 &51.2 &40.5 &35.6 &41.7 &30.7 &33.9 &39.1\\

MHFormer \cite{li2022mhformer}$\dagger$ f=243 &\textbf{31.5} &\textbf{34.9} &\textbf{32.8} &\textbf{33.6} &\textbf{35.3} &\underline{39.6} &\textbf{32.0} &\textbf{32.2} &\textbf{43.5} &\underline{48.7} &\textbf{36.4} &\underline{32.6} &\underline{34.3} &\textbf{23.9} &\textbf{25.1} &\textbf{34.4}\\
Pose-Oriented \cite{li2023pose} &-&-&-&-&-&-&-&-&-&-&-&-&-&-&- &-\\
diffPose \cite{gong2023diffpose} &\underline{33.9} &38.2 &36.0 &39.2 &40.2 &46.5 &35.8 &34.8 &48.0 &52.5 &41.2 &36.5 &40.9 &30.3 &33.8 &39.2\\
  \hline
Ours$*$	&34.4	&\underline{35.8}	&34.1	&\underline{34.9}	&\underline{36.4}	&\textbf{30.3}	&\underline{33.3}	&43.3	&\underline{48.0}	&\textbf{37.0}	&\underline{37.6} &\textbf{32.4}	&\textbf{28.5}	&36.9	&31.7	&\underline{36.0}\\

\hline
  \end{tabular}
}
\caption
  {
    Quantitative comparison with the state-of-the-art methods on Human3.6M under Protocol 2, taking SH (Stacked Hourglass network \cite{newell2016stacked}) 2D poses (top), taking CPN (Cascaded pyramid network) 2D poses (bottom) as inputs. 
    ($\dagger$) - uses temporal information.  $(*)$ - uses image information.
    \textbf{Blod}: best; 
    \underline{Underlined}: second best.  
  } 
  \label{appendix:pmpjpe}

\end{table*}

\begin{table}
  \centering
\renewcommand\arraystretch{1.1}
  \begin{tabular}{ccc}
   \hline
    Dataset & Task & MPJPE\\
    \hline
    ImageNet  & Image Classification & 35.2\\
    MS COCO &  2D human pose estimation & \textbf{28.6}\\ 
   \hline
  \end{tabular}
    \caption{
    Effect on the Image Encoder Pretraining.}
  \label{table:appendix1}
\end{table}

\vskip 0.05 in
\noindent\textbf{Additional ablation study on Pose-guided Transformer Layer} We conduct an additional ablation study to diagnose the effect of the Pose-guided dual attention. To evaluate the effect of ``Normalization before Transposition", we conduct an ablation study for another two types:  ``Normalization after Transposition" and ``double Normalization", which means first do Normalization, then do Transposition, then do Normalization again. Besides, we conducted experiments that employ two independent cross-attention (Image-to-keypoints cross-attention and keypoints-to-Image cross-attention) for comparison. The results are shown in Table \ref{table:appendix3}. ``Normalization before Transposition" achieves the best result both in H3.6M dataset and 3DHP dataset, which shows the effectiveness of our design.

\begin{table}
  \centering
\renewcommand\arraystretch{1.1}
  \begin{tabular}{cc|cc}
   \hline
    Backbone & MPJPE &  Backbone & MPJPE\\
   \hline
        Resnet50& \textbf{33.7} & HRNet-w32 &\textbf{28.6}\\
     Resnet101 & 37.5 &HRNet-w48 & 30.3\\
   \hline
  \end{tabular}
    \caption{
    Ablation study on different Image Encoders.}
  \label{table:appendix2}
\end{table}

\begin{table}
  \centering
\renewcommand\arraystretch{1.1}
  \begin{tabular}{c|cc}
   \hline
    Strategy & H3.6M & 3DHP\\
   \hline
    Normalization before Transposition &  \textbf{29.2} & \textbf{70.9}\\
    Normalization after Transposition & 29.6 & 80.9\\
    Double Normalization & 29.8 & 71.1\\
    Two Cross-attention & 30.2 & 72.8\\
   \hline
  \end{tabular}
    \caption{
    Effect on "Normalization before Transposition" strategy in Pose-guided Transformer Layer.}
  \label{table:appendix3}
\end{table}
\end{document}